\documentclass[conference]{IEEEtran}
\IEEEoverridecommandlockouts

\usepackage[pdftex]{graphicx}

\usepackage{times}
\usepackage{epsfig}
\usepackage{multirow}

\usepackage{cite}
\usepackage{amsmath,amssymb,amsfonts}
\usepackage{algorithmic}
\usepackage{graphicx}
\usepackage{textcomp}
\usepackage{xcolor}

\usepackage[pagebackref=true,breaklinks=true,colorlinks,bookmarks=false]{hyperref}

\def\BibTeX{{\rm B\kern-.05em{\sc i\kern-.025em b}\kern-.08em
    T\kern-.1667em\lower.7ex\hbox{E}\kern-.125emX}}

\begin{document}

\title{YOLO5Face: Why Reinventing a Face Detector}


\author{
\IEEEauthorblockN{Delong Qi, Weijun Tan*, Qi Yao, Jingfeng Liu }
\IEEEauthorblockA{\textit{Shenzhen Deepcam Information Technologies}\\Shenzhen, China \\
\{delong.qi,weijun.tan,qi.yao,jingfeng.liu\}@deepcam.com\\
*LinkSprite Technologies, USA, weijun.tan@linksprite.com}
}

\maketitle

\begin{abstract}
 Tremendous progress has been made on face detection in recent years using convolutional neural networks. While many face detectors use designs designated for the detection of face, we treat face detection as a general object detection task. We implement a face detector based on YOLOv5 object detector and call it YOLO5Face. We add a five-point landmark regression head into it and use the Wing loss function. We design detectors with different model sizes, from a large model to achieve the best performance, to a super small model for real-time detection on an embedded or mobile device. Experiment results on the WiderFace dataset show that our face detectors can achieve state-of-the-art performance in almost all the Easy, Medium, and Hard subsets, exceeding the more complex designated face detectors. The code is available at \url{https://www.github.com/deepcam-cn/yolov5-face}.  
 \end{abstract}

\begin{IEEEkeywords}
Face detection, convolutional neural network, YOLO, real-time, embedded device, object detection 
\end{IEEEkeywords}

\section{Introduction}

Face detection is a very important computer vision task. Tremendous progresses have been made since deep learning, particularly convolutional neural network (CNN), has been used in this task. As the first step of many tasks, including face recognition, verification, tracking, alignment, expression analysis, face detection attracts many researches and developments in the academia and the industry. And the performance of face detection has improved significantly over the years. For a survey of the face detection, please refer to the benchmark results \cite{WiderFace}, \cite{WiderFaceWeb}.   
There are many methods in this field from different perspectives. Research directions include design of CNN network, loss functions, data augmentations, and training strategies. For example, in the YOLOv4 paper, the authors explore all these research directions and propose the YOLOV4 object detector based on optimizations of network architecture, selection of bags of freebies, and selection of bags of specials \cite{YOLOv4}.  

In our approach, we treat the face detection as a general object detection task.  We have the same intuition as the TinaFace \cite{TinaFace}. Intuitively, face is an object. As discussed in the TinaFace \cite{TinaFace}, from the perspective of data, the properties that faces has, like pose, scale, occlusion, illumination, blur and etc., also exist in other objects. The unique properties in faces like expression and makeup can also correspond to distortion and color in objects.  Landmarks are special to face, but they are not unique either. They are just key points of an object. For example, in license plate detection, landmarks are also used. And adding landmark regression in the object prediction head is straightforward. Then from the perspective of challenges encountered by face detection like multi-scale, small faces and dense scenes, they all exist in generic object detection. Thus, face detection is just a sub task of general object detection. 

In this paper, we follow this intuition and design a face detector based on the YOLOv5 object detector \cite{YOLOv5}. We modify the design for face detection considering large faces, small faces, landmark supervision, for different complexities and applications. Our goal is to provide a portfolio of models for different applications, from very complex ones to get the best performance to very simple ones to get the best trade-off of performance and speed on embedded or mobile devices.  

Our main contributions are summarized as following, 
\begin{itemize}
\item {We redesign the YOLOV5 object detector \cite{YOLOv5} as a face detector, and call it YOLO5Face. We implement key modifications to the network to improve the performance in terms of mean average precision (mAP) and speed. The details of these modifications will be presented in Section III.}
\item {We design a series of models of different model sizes, from large models, to medium models, to super small models, for needs in different applications. In addition to the backbone used in YOLOv5 \cite{YOLOv5}, we implement a backbone based on ShuffleNetV2 \cite{ShuffleNetv2}, which gives the state-of-the-art (SOTA) performance and fast speed for mobile device.} 
\item {We evaluate our models on the WiderFace \cite{WiderFace} dataset. On VGA resolution images, almost all our models achieve the SOTA performance and fast speed. This proves our goal, as the tile of this paper claims, we do not need to reinvent a face detector since the YOLO5Face can accomplish it.}

\end{itemize}


\section{Related Work}

\subsection {Object Detection}

General object detection aims at locating and classifying the pre-defined objects in a given image. Before deep CNN is used, traditional face detection uses hand crafted features, like HAAR, HOG, LBP, SIFT, DPM, ACF, etc. The seminal work by Viola and Jones \cite{VJ} introduces integral image to compute HAAR-like features. For a survey of face detection using hand crafted features, please refer to \cite{survey1,survey2}.       

Since the deep CNN shows its power in many machine learning tasks, face detection is dominated by deep CNN methods. There are two-stage and one-stage object detectors. Typical two-stage methods are the RCNN family, including RCNN \cite{RCNN}, fast-RCNN \cite{fast-RCNN}, faster-RCNN \cite{faster-RCNN}, mask-RCNN \cite{mask-RCNN}, Cascade-RCNN \cite{cascade-RCNN}. 

The two-stage object detector have very good performance but suffers from long latency and slow speed. In order to overcome this problem, one-stage object detectors are studied. Typical one-stage networks include SSD \cite{SSD}, YOLO \cite{YOLOv1, YOLOv2, YOLOv3, YOLOv4, YOLOv5}. 

Other object detection networks include FPN \cite{FPN}, MMDetection \cite{MMDetection}, EfficientDet \cite{EfficientDet}, transformer(DETR) \cite{DETR}, Centernet \cite{CenterNet,CenterNet2}, and so on. 

\subsection {Face Detection}

The researches for face detection follows the general object detection.  After the most popular and challenging face detection benchmark WiderFace dataset \cite{WiderFace} is released, face detection develops rapidly focusing on the extreme and real variation problem including scale, pose, occlusion, expression, makeup, illumination, blur and etc.

A lot of methods are proposed to deal with these problems, particularly the scale, context, anchor in order to detect small faces.  These methods include MTCNN \cite{MTCNN}, FaceBox \cite{FaceBox}, S3FD \cite{S3FD}, DSFD \cite{DSFD}, RetinaFace \cite{RetinaFace}, RefineFace \cite{RefineFace}, and the most recent ASFD \cite{ASFD}, MaskFace \cite{MaskFace}, TinaFace \cite{TinaFace}, MogFace \cite{MogFace}, and SCRFD \cite{SCRFD}. For a list of popular face detectors, the readers are referred to the WiderFace website \cite{WiderFaceWeb}.    

It is worth noting that some of these face detectors explore unique characteristics in human face, the others are just general object detector adopted and modified for face detection. Use RetinaFace \cite{RetinaFace} as an example, it uses landmark (2D and 3D) regression to help the supervision of face detection, while TinaFace \cite{TinaFace} is simply a general object detector. 

\subsection {YOLO}

YOLO first appeared in 2015 \cite{YOLOv1} as a different approach than popular two-stage approaches. It treats object detection as an regression problem rather than a classification problem.  It performs all the essential stages to detect an object using a single neural network. As a result, it not only achieves very good detection performance, but also achieves real-time speed. Furthermore, it has excellent generalization capability, can be easily trained to detect different objects. 

Over the next five years, the YOLO algorithm have been upgraded to five versions with many innovative ideas from the object detection community. The first three versions - YOLOv1 \cite{YOLOv1},YOLOv2 \cite{YOLOv2}, YOLOv3 \cite{YOLOv3}are developed by the author of the original YOLO algorithm. Out of these three versions, the YOLOv3 \cite{YOLOv3} is a milestone with big improvements in performance and speed by introducing multi-scale features (FPN) \cite{FPN}, better backbone network (Darknet53), and replacing the Softmax classification loss with the binary cross-entropy loss.  

In early 2020, after the original YOLO authors withdrawn from the research field, YOLOv4 \cite{YOLOv4} was released by a different research team. The team explore a lot of options in almost all aspects of the YOLOv3 \cite{YOLOv3} algorithm, including the backbone, and what they call bags of freebies, and bags of specials. It achieves 43.5\% AP (65.7\% AP50) for the MS COCO dataset at a real time speed of 65 FPS on Tesla V100.  

One month later, the YOLOv5 \cite{YOLOv5} was released by another different research team. In the algorithm prospective, the YOLOv5 \cite{YOLOv5} does not have many innovations. And the team does not publish a paper. These bring quite some controversies about if it should be called YOLOv5. However, due to its significantly reduced model size, faster speed, and similar performance as YOLOv4 \cite{YOLOv4}, and a full implementation in Python (Pytorch), it is welcome by the object detection community. 

\section{YOLO5Face Face Detector}

In this section we present the key modifications we make in YOLOv5 and make it a face detector - YOLO5Face.   

\subsection{Network Architecture}

We use the YOLOv5 object detector \cite{YOLOv5} as our baseline and optimize it for face detection. We introduce some modifications designated for detection of small faces as well as large faces.  

\begin{figure*}
    \centering
    \includegraphics[scale=0.670]{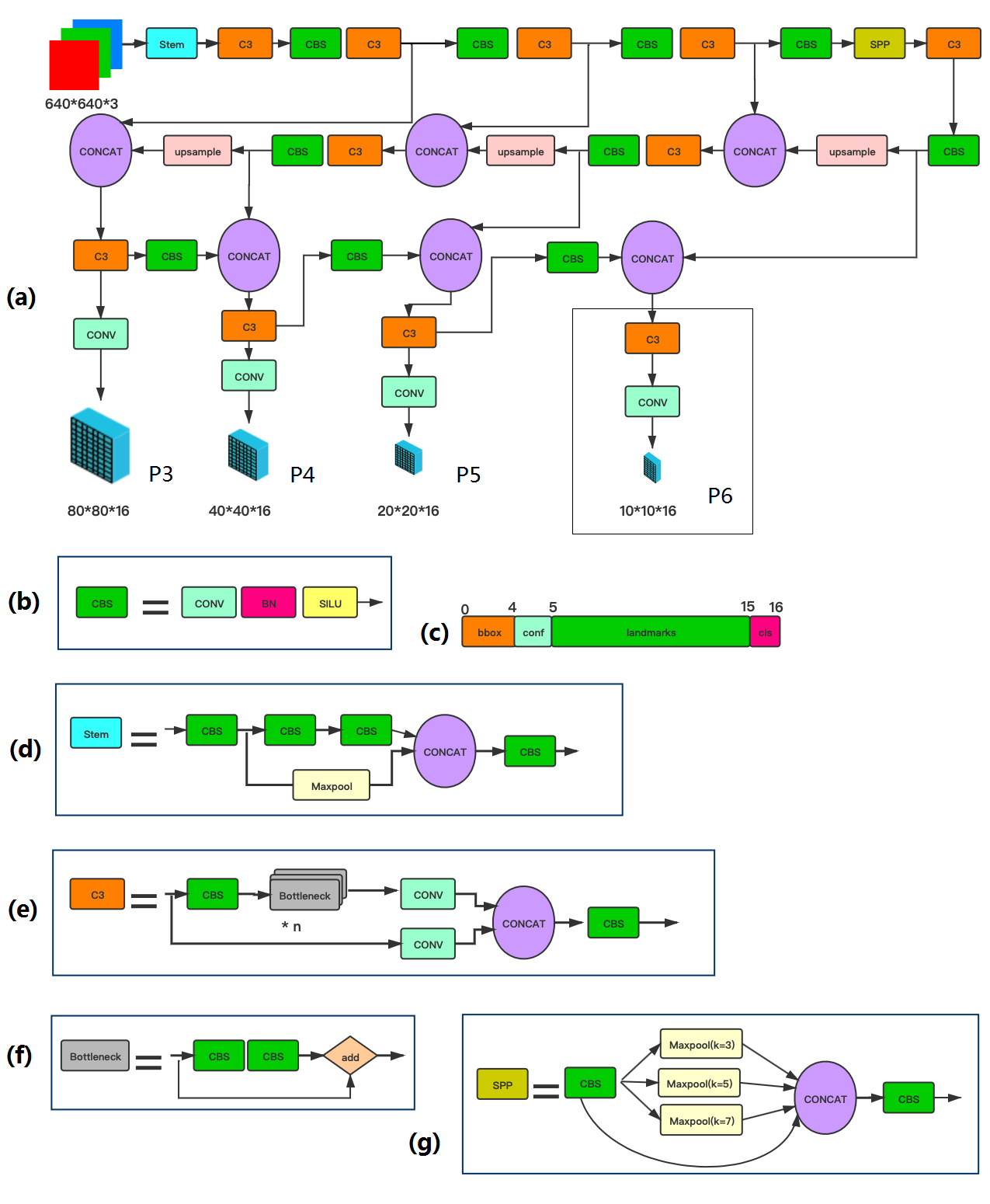}
    \caption{The proposed YOLO5Face network architecture.}
    \label{network}
\end{figure*}

The network architecture of our YOLO5Face face detector is depicted in Fig. \ref{network}. It consists of the backbone, neck, and head. In YOLOv5, a new designed backbone called CSPNet \cite{YOLOv5} is used. In the neck, an SPP \cite{SPP} and a PAN \cite{PAN} are used to aggregate the features.  In the head, regression and classification are both used. 

In Fig. \ref{network} (a), the overall network architecture is depicted. In Fig. \ref{network} (b), a key block called CBS is defined, which consists of Conv layer, BN layer, and a SILU \cite{SILU} activation function. This CBS block is used in many other blocks. In Fig. \ref{network} (c), an output label for the head is shown, which include bounding box (bbox), confidence (conf), classification (cls) and five-point landmarks. The landmarks are our addition to the YOLOv5 to make it a face detector with landmark output. If without the landmark, the last dimension 16 should be 6. Please note that, the output dimensions 80*80*16 in P3, 40*40*16 in P4, 20*20*16 in P5, 10*10*16 in optional P6 are for every anchor. The the real dimension should be multiplied by the number of anchors.  

In Fig. \ref{network} (d), a Stem structure \cite{Stem} is shown, which is used to replace the original Focus layer in YOLOv5. The introduction of the Stem block into YOLOv5 for face detection is one of our innovations. 

In Fig. \ref{network} (e), a CSP block (C3) is shown. This block is inspired by the DenseNet \cite{DenseNet}. However, instead of adding the full input and the output after some CNN layers, the input is separated two two halves. One half is passed through a CBS block, a number of Bottleneck blocks, which is shown in Fig. \ref{network} (f), then another Conv layer. The other half is passed through a Conv layer, then the two are concatenated, followed by another CBS block.     

Fig. \ref{network} (g), an SPP block \cite{SPP} is shown. In this block the three kernel sizes 13x13, 9x9, 5x5 in YOLOv5 are revised to 7x7, 5x5, 3x3 in our face detector.  This has been shown as one of the innovations that improves the face detection performance. 

Note that we only consider VGA resolution input images. To be more precise, the longer edge of the input image is scaled to 640, and the shorter edge is scaled accordingly. The shorter edge is also adjusted to be a multiple of the largest stride of the SPP block. For example, when P6 is not used, the shorter edge needs to be multiple of 32; when P6 is used, the shorter edge needs to multiple of 64.   

\subsection{Summary of Key Modifications}

The key modifications are summarized as follows.  

\begin{itemize}
\item {We add a landmark regression head to the YOLOv5 network. The Wing loss \cite{wingloss} is used a loss function for it. This makes the face detector more useful since landmarks are used in many applications. The landmark locations are more accurate. This extra supervision helps the face detector accuracy.}
\item {We replace the Focus layer of YOLOv5 \cite{YOLOv5} with a Stem block structure \cite{Stem}. It increases the network's generalization capability, and reduces the computation complexity while the performance does not degrade.} 
\item {We change the SPP block \cite{SPP} and use a smaller kernel. It makes the YOLOv5 more suitable for face detection and improve the detection accuracy.} 
\item {We add a P6 output block with stride of 64. It increases the capability to detect large faces. This is an item easily overlooked by many researchers since their focuses are more on the detection of small faces.}
\item {We find that some data augmentation methods on general object detection are not appropriate on face detection, including up-down flipping and Mosaic. Removing the up-down flipping improves the performance. When small images are used, the Mosaic augmentation \cite{YOLOv4} degrades the performance. However, when the small faces are ignored, it works well. Random cropping helps the performance.}
\item {We design two super light-weight models based on ShuffleNetV2 \cite{ShuffleNetv2}. This backbone is very different from the CSP network. These models are super small, while achieve SOTA performance for embedded or mobile device.}
\end{itemize}

\subsection{Landmark Regression}

Landmarks are important characteristics for human face. They can be used to do face alignment, face recognition, face express analysis, age analysis etc. Traditional landmarks consist of 68 points. They are simplified to 5 points in MTCNN \cite{MTCNN} Since then, the five-point landmarks have been used widely in face recognition. The quality of landmarks affects the quality of face alignment and face recognition. 

The general object detector does not include landmarks. It is straightforward to add it as a regression head. Therefore, we add it into our YOLO5Face. The landmark outputs will be used in align face images before they are sent to the face recognition network. 

General loss functions for landmark regression are L2, L1, or smooth-L1. The MTCNN \cite{MTCNN} uses the L2 loss function.  However, it is found these loss functions are not sensitive to small errors. To overcome this problem, the Wing-loss is proposed \cite{wingloss},      

\begin{equation}
    wing(x)= 
\begin{cases}
    w \cdot ln(1+|x|/e) ,& \text{if } x < w\\
    |x|-C,              & \text{otherwise}
\end{cases}
\label{eq1}
\end{equation}
The non-negative $w$ sets the range of the nonlinear part to $(-w, w)$, $e$ limits the curvature of the nonlinear region and $C=w-w ln(1+w/e)$ is a constant that smoothly links the piecewise-defined linear and nonlinear parts. Plotted in Fig. 2 is this Wing loss function with different parameters $w$and $e$ It can be seen that the response at small error area near zero is boosted compared to the L2, L1, or smooth-L1 functions.   

The loss functions for landmark point vector  $s=\{s_i\}$, and its ground truth $s'=\{s_i\}$, where $i=1,2,...,10$, is defined as, 
\begin{equation}
    loss_L(s) = \sum_{i}{wing(s_i-s'_i)}
\label{eq2}
\end{equation}
Let the general object detection loss function of YOLOv5 be $loss_O(bounding\_box,class,probability)$, then the new total loss function is, 
\begin{equation}
    loss(s) = loss_O + \lambda_L\cdot loss_L
\label{eq3}
\end{equation}
where the $\lambda_L$ is a weighting factor for the landmark regression loss function. 

\subsection{Stem Block Structure}

We use a stem block similar to \cite{Stem}. The stem block is shown in Fig.\ref{network} (d). With this stem block, we implement a stride = 2  in the first spatial down-sampling on the input image, and increase the number of channels. With this stem block, the computation complexity only increase marginally, while a strong representation capability is ensured.  


\begin{table}[htb]
    \centering
    \begin{tabular}{c|c|c|c}
        \hline
        Model &  Backbone & (D,W) & With P6? \\
        \hline
        YOLOv5s	& YOLO5-CSPNet \cite{YOLOv5} & (0.33,0.50) & No \\
        YOLOv5s6	& YOLO5-CSPNet & (0.33,0.50) & Yes \\ 
        YOLOv5m	& YOLO5-CSPNet & (0.50,0.75) & No \\
        YOLOv5m6	& YOLO5-CSPNet & (0.50,0.75) & Yes \\
        YOLOv5l	& YOLO5-CSPNet & (1.0,1.0) & No \\
        YOLOv5l6	& YOLO5-CSPNet & (1.0,1.0) & Yes \\
        \hline
        YOLOv5n	& ShuffleNetv2 \cite{ShuffleNetv2} & - & No \\
        YOLOv5n-0.5	& ShuffleNetv2-0.5 \cite{ShuffleNetv2} & - & No \\ 	

        \hline
    \end{tabular}
    \caption{Detail of implemented YOLO5Face models, where (D,W) are the depth and width multiples of the YOLOv5 CSPNet \cite{YOLOv5}. The number of parameters and Flops are listed in Table \ref{t2}.}
    \label{t0}
\end{table}

\begin{table*}[!htb]
    \centering
    \begin{tabular}{c|c|c|c|c|c|c}
        \hline
        Modification & Method & Easy & Medium &  Hard & Params(M) & Flops(G)  \\
        \hline

        \multirow{2}{*}{Stem block} & Focus+Conv & 93.56 & 92.54 & 82.56 & 7.091 & 6.174 \\ \cline{2-7}
        &  Stem Block & 94.13 & 92.87 & 82.79 & 7.075 &5.751 \\
        \hline
        \multirow{2}{*}{SPP Kernel} & (13,9,5) & 93.43 & 91.12 & 82.64 & - & - \\ \cline{2-7}
        & (7,5,3) & 94.33 & 92.61 & 84.15 & - & - \\
        \hline
        \multirow{2}{*}{P6 block} & No & 94.31  & 92.52 & 83.15 & 7.075 &5.751 \\ \cline{2-7}
         & Yes & 95.29 & 93.61 & 83.13 & 12.386 & 6.28 \\
        \hline
        \multirow{4}{*}{Data augmentation}  & Baseline (with Mosaic) & 91.34  & 90.21 & 83.54 & - & - \\ \cline{2-7}
         & - up-down flipping & 91.87 & 90.56 & 83.58 & - & - \\ \cline{2-7}
         & + Ignore small faces & 94.12 & 92.21 & 82.21 & - & - \\ \cline{2-7}
         & + Random crop & 94.34 & 92.58 & 83.17 & - & - \\ \cline{2-7}
        \hline
    \end{tabular}
    \caption{Ablation study results on the WiderFace validation dataset.}
    \label{t1}
\end{table*}

\begin{table*}[!htb]
    \centering
    \begin{tabular}{c|c|c|c|c|c|c}
        \hline
        Detector & Backbone & Easy & Medium &  Hard & Params(M) & Flops(G)  \\
        \hline
       DSFD \cite{DSFD} & ResNet152 \cite{Resnet} & 94.29  & 91.47 &  71.39 &  120.06  & 259.55 \\
       RetinaFace \cite{RetinaFace}  & ResNet50 \cite{Resnet}  & 94.92 &  91.90 &  64.17 &  29.50 &  37.59\\
       HAMBox \cite{HAMBox}  &  ResNet50 \cite{Resnet} &  95.27 &  93.76 &  76.75 &  30.24 &  43.28\\
       TinaFace \cite{TinaFace} &  ResNet50 \cite{Resnet}  & 95.61 &  94.25 &  81.43 &  37.98 &  172.95\\
       SCRFD-34GF \cite{SCRFD} &  Bottleneck ResNet  & \textbf{96.06} &  \textbf{94.92} &  \textbf{85.29} &  9.80 &  34.13\\
       SCRFD-10GF \cite{SCRFD} &  Basic ResNet \cite{Resnet} &  95.16  & 93.87 &  83.05 &  3.86 &  9.98\\
       
       \textbf{Our YOLOv5s} & YOLOv5-CSPNet \cite{YOLOv5} &94.33 & 92.61 & 83.15 & 7.075 & 5.751 \\
       \textbf{Our YOLOv5s6} & YOLOv5-CSPNet & 95.48 & 93.66 & 82.8 & 12.386 & 6.280 \\
       \textbf{Our YOLOv5m} & YOLOv5-CSPNet & 95.30 & 93.76 & 85.28 & 21.063 & 18.146 \\
       \textbf{Our YOLOv5m6} & YOLOv5-CSPNet & 95.66 & 94.1 & 85.2 & 35.485 & 19.773 \\
       \textbf{Our YOLOv5l} & YOLOv5-CSPNet & 95.9 & 94.4 & 84.5 & 46.627 & 41.607 \\
       \textbf{Our YOLOv5l6} & YOLOv5-CSPNet & 96.38 & 94.90 & 85.88 & 76.674 & 45.279 \\
       \textbf{Our YOLOv5x6} & YOLOv5-CSPNet & \textbf{96.67} & \textbf{95.08} & \textbf{86.55} & 141.158 & 88.665 \\
       
       \hline
       SCRFD-2.5GF \cite{SCRFD} &  Basic Resnet  & \textbf{93.78} &  \textbf{92.16} &  \textbf{77.87} &  0.67 &  2.53\\ 
       SCRFD-0.5GF \cite{SCRFD}	&Depth-wise Conv& 90.57&88.12&	68.51&	0.57	&0.508 \\
       RetinaFace \cite{RetinaFace} &	MobileNet0.25\cite{MobileNetV2}	&87.78	&81.16&	47.32&	0.44&	0.802\\
       FaceBoxes \cite{FaceBox} &	- &	76.17&	57.17&	24.18&	1.01&	0.275\\
       
       \textbf{Our YOLOv5n} & ShuffleNetv2 \cite{ShuffleNetv2} & \textbf{93.61} & \textbf{91.54} & \textbf{80.53} & 1.726 & 2.111 \\  
       \textbf{Our YOLOv5n0.5} & ShuffleNetv2-0.5 \cite{ShuffleNetv2} & 90.76&88.12&73.82& 0.447 & 0.571 \\ 
       \hline
       \end{tabular}
    \caption{Comparison of our YOLO5Face and existing face detectors on the WiderFace validation dataset \cite{WiderFace}.}
    \label{t2}
\end{table*}

\subsection{SPP with Smaller Kernels}

Before forwarding to feature aggregation block in the neck, the output feature maps of the YOLO5 backbone are sent to an additional SPP block \cite{SPP} to increase the receptive field and separate out the most important features. Instead of many CNN models containing fully connected layers which only accept input images of specific dimensions, SPP is proposed to aim at generating a fixed-size output irrespective of the input size. In addition, SPP also helps to extract important features by pooling multi-scale versions of itself.  

In YOLO5, three kernel sizes 13x13,9x9,5x5 are used \cite{YOLOv5}. We revise them to use smaller size kernels 7x7, 5x5 and 3x3. These smaller kernels help to detect small faces more easily, and increase the overall face detection performance.     

\subsection{P6 Output Block}

The backbone of YOLO object detector has many layers. As the feature becomes more and more abstract as the layers go deeper, 
the spatial resolution of feature maps decreases due to downsampling, which leads to to a loss of spatial information as well as fine-grained features. In order to preserve these fine-grained features, the FPN \cite{FPN} is introduced to YOLOv3 \cite{YOLOv3}.  

In FPN \cite{FPN}, the fine-grained features take a long path traveling from low-level to high-level layers. To overcome this problem, the PAN is proposed to add a bottom-up augmentation path along the top-down path used in FPN. In addition, in the connection of the feature maps to the lateral architecture, the element-wise addition operation is replaced with concatenation. In FPN, object predictions are done independently on different scale levels, which do not utilize information from other feature maps, and may produce duplicated predictions. In PAN \cite{PAN},  the output feature maps of bottom-up augmentation pyramid are fused by using (Region of Interest) ROI align and fully connected layers with element-wise max operation. 

In YOLOv5, there are three output blocks in the PAN output feature maps, called P3,P4,P5 corresponding to 80x80x16, 40x40x16, 20x20x16, with strides 8,16,32, respectively. In our YOLO5Face, we add an extra P6 output block, whose feature map is 10x10x16 with stride 64. This modification particularly helps the detection of large faces. While almost all face detectors focus on improving detection of small faces, detection of large faces can be easily overlooked. We fill this hole by adding the P6 output block.  

\subsection{ShuffleNetV2 as Backbone}

The ShuffleNet \cite{ShuffleNet} is an extremely efficient CNN for mobile device. The key block is called the ShuffleNet block.  It utilizes two new operations, pointwise group convolution and channel shuffle, to greatly reduce computation cost while maintaining accuracy. 

The ShuffleNetv2 \cite{ShuffleNet} is an improved version of ShuffleNet. It borrows the shortcut network architecture similar to the DenseNet \cite{DenseNet}, and the the element wise addition is changed to concatenation, similar to the change in PAN \cite{PAN} in YOLOv5 \cite{YOLOv5}. But different from DenseNet, ShuffleNetV2 does not densely concatenate, and after the concatenation, the channel shuffling is used to mix the features.  This makes the ShuffleNetV2 a super fast network. 

We use the ShuffleNetV2 as the backbone in YOLOv5 and implement super small face detectors YOLOv5n-Face, and YOLOv5n0.5-Face. 

\section{Experiments}

\subsection{Dataset}

The WiderFace dataset \cite{WiderFace} is the largest face detection dataset, which contains 32,203 images and 393,703 faces.  For its large variety of scale, pose, occlusion, expression, illumination and event, it is close to reality and is very challenging.  

The whole dataset is divided into train/validation/test sets by ratio 50\%/10\%/40\% within each event class. Furthermore, each subset is defined into three levels of difficulty: Easy, Medium, and Hard. As it names indicates, the Hard subset is most challenging. So the performance on the Hard subset reflects best the effectiveness of a face detector.  

Unless specified otherwise, the WiderFace dataset \cite{WiderFace} is used in this work. In the face recognition with YOLO5Face landmark and alignment, the Webface dataset \cite{webface} is used. The FDDB dataset \cite{fddb} is used in testing to demonstrate our model's performance on cross-domain datasets.   

\subsection{Implementation Details}

We use the YOLOv5-4.0 codebase \cite{YOLOv5} as our starting point and implement all the modifications we describe earlier in PyTorch.  

The SGD optimizer is used. The initial learning rate is 1E-2, the final learning rate is 1E-5, and the weight decay is 5E-3. A momentum of 0.8 is used in the first three warming-up epochs. After that, the momentum is changed to 0.937. The training runs 250 epochs with a batch size of 64. The ${\lambda_L=0.5}$ is optimized by exhaust search.     

\begin{figure*}
    \centering
    \includegraphics[width=17cm]{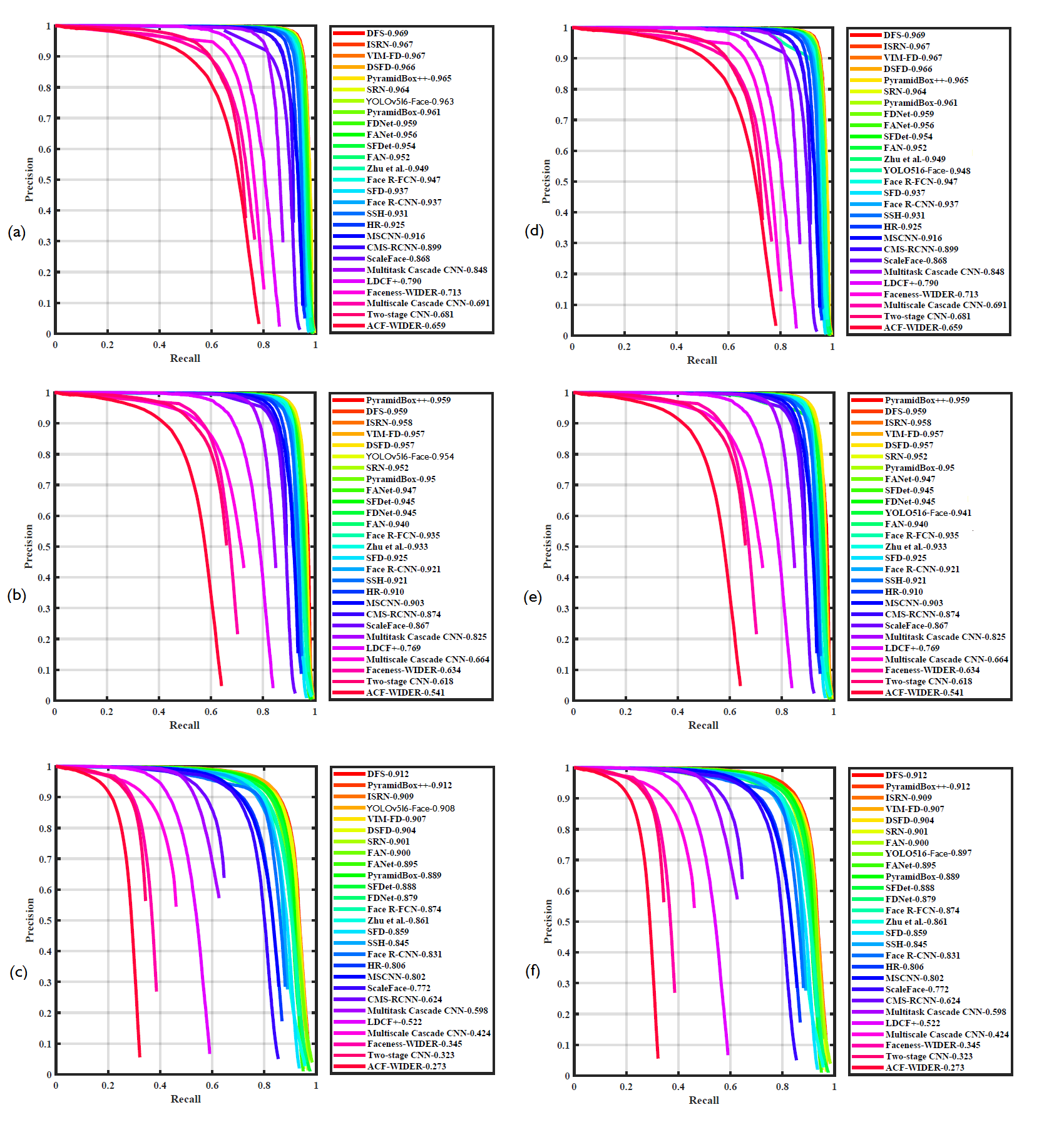}
    \caption{The precision-recall (PR) curves of face detectors, (a) validation-Easy, (b)validation-Medium, (c) validation-Hard, (d) test-Easy, (e) test-Medium, (f) test-Hard. }
    \label{pr}
\end{figure*}

\textbf{Implemented Models}. We implement a series of face detector models, as listed in Table \ref{t0}. We implement eight relatively large models, including extra large-size models (YOLOv5x, YOLOv5x6), large-size models (YOLOv5l, YOLOv5l6) medium-size models (YOLOv5m, YOLOv5m6), and small-size models (YOLOv5s, YOLOv5s6). In the name of the model, the last postfix 6 means it has the P6 output block in the SPP. These models all use the YOLOv4 CSPNet as the backbone with different depth and width multiples, denoted as D and W in Table \ref{t0}. 

Furthermore, we implement two super small-size models, YOLOv5n and YOLOv5n0.5, which use the ShuffleNetv2 and ShuffleNetv2-0.5 \cite{ShuffleNetv2} as the backbone. Except for the backbone, all other main blocks, including the stem block, SPP, PAN, are the same as in the larger models. 

The number of parameters and number of flops of all these models is listed in Table \ref{t2} for comparison with existing methods. 

\subsection{Ablation Study}
In this subsection we present the effects of the modifications we have in our YOLO5Face. In this study we use the YOLO5s model. We use the WiderFace \cite{WiderFace} validation dataset and use the mAP as the performance metric.

\textbf{Stem Block vs. Focus Layer.} The mAP performances of the stem block \cite{Stem} and the focus layer are listed in first panel of Table \ref{t1}. Also listed are the number of parameters and number of flops. From the results we see that the stem block improves the mAP by 0.57\%, 0.33\%, and 0.23\% on the easy, medium, and hard subset, respectively.  

\textbf{SPP with Smaller Size Kernels.} The mAP performances of the SPP \cite{SPP} kernel sizes (7x7,5x5,3x3) and original kernel sizes (13x13,9x9,5x5) are listed in the second panel of Table \ref{t1}. From the results we see that the smaller kernel sizes improve the mAP by 0.9\%, 1.49\%, and 1.41\% on the easy, medium, and hard subset, respectively. The improvements are larger than that from the Stem block \cite{Stem}.   

\textbf{P6 Output Block.} The mAP performances of the addition of the P6 output block are listed in the third panel of Table \ref{t1}. From the results we see that the P6 block improves the mAP by 0.98\%, 1.09\%, and -0.02\% on the easy, medium, and hard subset, respectively.

\textbf{Data Augmentation} Performance results of a few data augmentation methods are listed in the fourth panel of Table \ref{t1}. From the results we see that ignoring small faces, random crop help the mAP in the Easy and Medium dataset, while the Mosaic \cite{YOLOv4} helps the mAP in the Hard dataset. As we explain before, the Mosaic has to work with the ignoring small faces, otherwise the performance degrades dramatically.

Please note that in these experiments the network configurations are not incremental. However in each of set of experiment, the baselines for the two networks are the same to make the comparison fair. For example in the SPP experiments, except for the kernel sizes are different, all other settting are identical.  

\begin{table}[tb]
    \centering
    \begin{tabular}{c|c|c}
        \hline
        FaceDetect & traning dataset & FNMR\\
        \hline
        RetinaFace\cite{RetinaFace} & WiderFace \cite{WiderFace} & 0.1065 \\
        YOLOv5s &  WiderFace & 0.1060\\
        YOLOv5s &  +Multi-task facial\cite{Multi-task-facial} & 0.1058\\        
        YOLOv5m & WiderFace & 0.1056\\
        YOLOv5m & +Multi-task facial & \textbf{0.1051}\\        
        \hline
    \end{tabular}
    \caption{Evaluation of YOLO5Face landmark on face recognition on the Webface test dataset \cite{webface}.}
    \label{t3}
\end{table}

\subsection{YOLO5Face for Face Recognition}

Landmark is critical for face recognition accuracy. In RetinaFace \cite{RetinaFace}, the accuracy of the landmark is evaluated with the MSE between estimated landmark coordinates and their ground truth and with the face recognition accuracy. The results show that the RetinaFace has better landmarks than the older MTCNN \cite{MTCNN}. 

In this work, we also use face recognition to evaluate the accuracy of landmarks of the YOLO5Face. We use the Webface test dataset, which is the largest face dataset with noisy 4M identities/260M faces, and cleaned 2M identities/42M faces \cite{webface}. This dataset is used in the ICCV2021 Masked Face Recognition (MFR) challenge \cite{ICCV21MFR}. In this challenge, both masked face images and standard face images are included, and a metric False Non-Match Rate (FNMR) at False Match Rate (FMR) = 1E-5 is used. The FNMR*0.25 for MFR plus FNMR*0.75 for standard face recognition is combined as the final metric. 

By default, the RetinaFace \cite{ResNet-D} is used as the face detector on the dataset. We compare our YOLO5Face with the RetinaFace on this dataset. We use ArcFace \cite{ArcFace} framework with Resnet124 \cite{Resnet} as backbone. Extracted features of two models trained on the Glint360k dataset \cite{glint360} are concatenated as the baseline model. We replace the  RetinaFace with our YOLO5Face. We test two models, a small model YOLOv5s, and a medium model YOLOv5m. More details can be found in \cite{ourMFR}. 

\begin{figure}[bt]
    \centering
    \includegraphics[width=8.25cm]{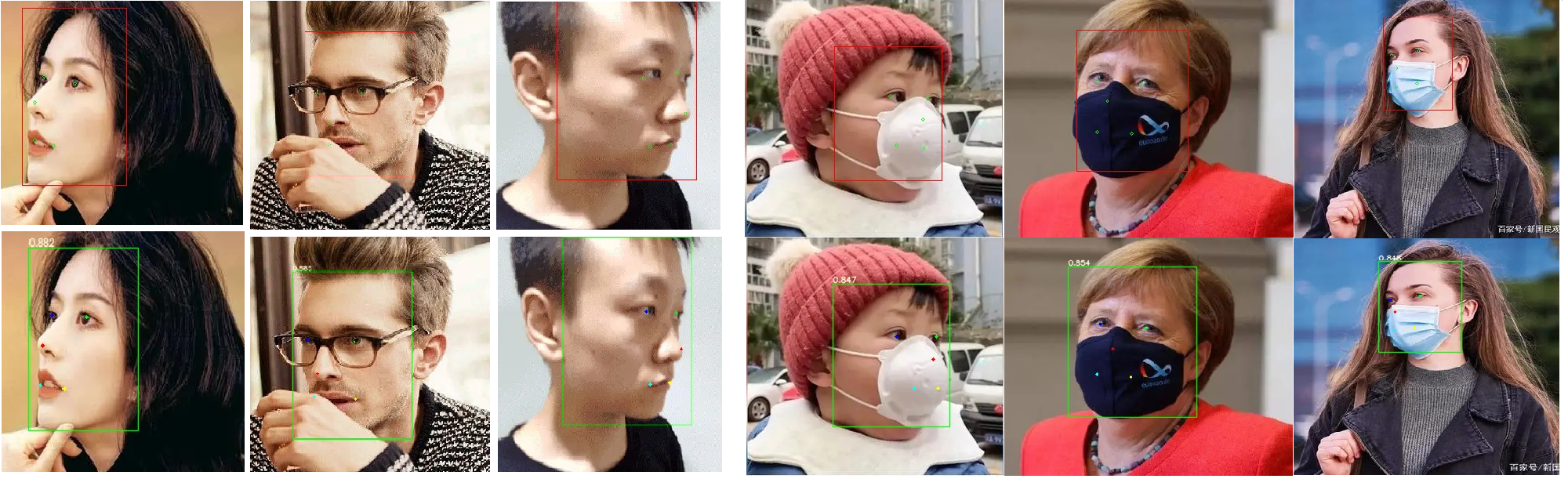}
    \caption{Some examples of detected face and landmarks, where the first row is from RetinaFace \cite{RetinaFace}, and second row is from our YOLOv5m.}
    \label{landmark}
\end{figure}

The results are listed in Table \ref{t3}. From the results, we see that both our small and medium models outperform the RetinaFace \cite{RetinaFace}. In addition, we notice that there are very few large face images in the WiderFace dataset, so we add some large face images from the Multi-task-facial dataset \cite{Multi-task-facial} into the YOLO5Face training dataset. We find that this technique improves face recognition performance.  

shown in Figure \ref{landmark} are some detected Webface \cite{webface} faces and landmarks using the RetinaFace \cite{RetinaFace} and our YOLOv5m. On the faces of a large pose, we can visually observe that our landmarks are more accurate, which has been prooved in our face recognition results shown in Table \ref{t3}.  

\subsection{YOLO5Face on WiderFace Dataset}

We compare our YOLO5Face with many existing face detectors on the WiderFace dataset. The results are listed in Table \ref{t2}, where the previous SOTA results and our best results are both highlighted.  

We first look at the performance of relatively large models whose number of parameters is larger than 3M and the number of flops is larger than 5G.  All existing methods achieve mAP in 94.27-96.06\% on the Easy subset, 91.9-94.92\% on the Medium subset, and 71.39-85.29\% on the Hard subset. The most recently released SCRFD \cite{SCRFD} achieves the best performance in all subsets. Our YOLO5Face (YOLOv5x6) achieves 96.67\%, 95.08\%, 86.55\% on the three subsets, respectively. We achieve the SOTA performance on all the Easy, Medium, and Hard subsets.  

Next, we look at the performance of super small models whose number of parameters is less than 2M and the number of flops is less than 3G. All existing methods achieve mAP in 76.17-93.78\% on the Easy subset, 57.17-92.16\% on the Medium subset, and 24.18-77.87\% on the Hard subset. Again, the SCRFD \cite{SCRFD} achieves the best performance in all subsets. Our YOLO5Face (YOLOv5n) achieves 93.61\%, 91.54\%, 80.53\% on the three subsets, respectively. Our face detector has a little bit worse performance than the SCRFD \cite{SCRFD} on the Easy and Medium subsets. However, on the Hard subset, our face detector is leading by 2.66\%. Furthermore, our smallest model, YOLOv5n0.5, has good performance, even its model size is much smaller.   

The precision-recall (PR) curves of our YOLO5Face face detector, along with the competitors, are shown in Figure \ref{pr}. The leading competitors include DFS \cite{DFS}, ISRN \cite{ISRN}, VIM-FD \cite{VIM-FD}, DSFD \cite{DSFD}, PyramidBox++ \cite{PyramidBox++}, SRN \cite{SRN}, PyramidBox \cite{PyramidBox} and more. For a full list of the competitors and their results on the WiderFace \cite{WiderFace} validation and test datasets, please refer to \cite{WiderFaceWeb}. In the results on the validation dataset, our YOLOv5x6-Face detector achieves 96.9\%, 96.0\%, 91.6\% mAP on the Easy, Medium, and Hard subset, respectively, exceeding the previous SOTA by 0.0\%, 0.1\%, 0.4\%. In the results on the test dataset, our YOLOv5x6-Face detector achieves 95.8\%, 94.9\%, 90.5\% mAP on the Easy, Medium, and Hard subset, respectively with 1.1\%, 1.0\%, 0.7\% gap to the previous SOTA. Please note that, in these evaluations, we only use multiple scales and left-right flipping without using other test-time augmentation (TTA) methods. Our focus is more on the VGA input images, where we achieve the SOTA in almost all conditions. 

\begin{table}[tb]
    \centering
    \begin{tabular}{c|c}
        \hline
        Method & MAP  \\
        \hline
        ASFD \cite{ASFD} & \textbf{0.9911} \\
        RefineFace \cite{RefineFace} & \textbf{0.9911} \\  
        PyramidBox \cite{PyramidBox} & 0.9869 \\
        FaceBoxes \cite{FaceBox} & 0.9598 \\ 
        Our YOLOv5s & 0.9843 \\
        Our YOLOv5m & 0.9849 \\
        Our YOLOv5l & 0.9867 \\
        Our YOLOv5l6 & 0.9880 \\
        \hline
    \end{tabular}
    \caption{Evaluation of YOLO5Face on the FDDB dataset \cite{fddb}.}
    \label{t4}
\end{table}

\subsection{YOLO5Face on FDDB Dataset}
FDDB dataset \cite{fddb} is a small dataset with 5171 faces annotated in 2845 images. To demonstrate our YOLO5Face's performance on the cross-domain dataset, we test it on the FDDB dataset without retraining on it. The performances of true positive rate (TPR) when the number of false-positive is 1000 are listed in Table 4. Please note that it is pointed out in RefineFace \cite{RefineFace} that the annotation of FDDB misses many faces. In order to achieve their performance of 0.9911, the RefineFace modifies the FDDB annotation. In our evaluation, we use the original FDDB annotation without modifications. RetinaFace \cite{RetinaFace} is not evaluated on the FDDB dataset.        


\section{Conclusion}
In this paper we present our YOLO5Face based on YOLOv5 object detector \cite{YOLOv5}. We implement eight models. Both the largest model YOLOv5l6 and the super small model YOLOv5n achieve close to or exceeding SOTA performance on the WiderFace \cite{WiderFace} validation Easy, Medium and Hard subsets. This proves the effectiveness of our YOLO5Face in not only achieving the best performance, but also running fast. Since we open-source the code, a lot of applications and mobile apps have been developed based on our design, and achieve impressive performance. 

\bibliographystyle{IEEEbib}
\bibliography{Yolo5Face}

\end{document}